\pdfoutput=1
\documentclass[11pt,a4paper]{article}
\usepackage{CJKutf8}

\usepackage{tcolorbox}
\usepackage{booktabs}
\usepackage{multirow}
\usepackage{array,graphicx}
\usepackage{listings}

\usepackage{times,latexsym}
\usepackage{url}
\usepackage[T1]{fontenc}

\usepackage[acceptedWithA]{tacl2021v1}
\usepackage[]{tacl2021v1}
\usepackage{xspace,mfirstuc,tabulary}

\newif\iftaclinstructions
\taclinstructionsfalse 
\iftaclinstructions
\renewcommand{\confidential}{}
\renewcommand{\anonsubtext}{(No author info supplied here, for consistency with
TACL-submission anonymization requirements)}
\newcommand{\instr}
\fi

\iftaclpubformat 

\else

\fi

\title{Revisiting Acceptability Judgements: \\ 
CoLAC - Corpus of Linguistic Acceptability in Chinese}

\author{
  Hai Hu\Thanks{First two authors contributed equally to this work. Correspondence: Hai Hu (hu.hai@sjtu.edu.cn) and Rui Wang (wangrui12@sjtu.edu.cn). } 
  \\
  {\footnotesize School of Foreign Languages}
  \\
  {\footnotesize Shanghai Jiao Tong University}
  \And
  Ziyin Zhang$^*$ 
    \\
  {\footnotesize Dept.~of Computer Science}
  \\
  {\footnotesize and Engineering}
  \\
  {\footnotesize Shanghai Jiao Tong University}
  \And
  Weifang Huang
  \\
  {\footnotesize School of Foreign Languages}
  \\
  {\footnotesize Shanghai Jiao Tong University}
  \AND
  Jackie Yan-ki Lai
  \\
  {\footnotesize Dept.~of Linguistics and Translation}
  \\
  {\footnotesize City University of Hong Kong}
  \And
  Aini Li
  \\
  {\footnotesize Dept.~of Linguistics}
  \\
  {\footnotesize University of Pennsylvania}
  \And
  Yina Patterson
\\
{\footnotesize Dept.~of Asian and Near Eastern Languages}
  \\
  {\footnotesize Brigham Young University}
  \AND
  Jiahui Huang
\\
{\footnotesize Dept.~of Linguistics}
  \\
  {\footnotesize University of Washington}
  \And
  Peng Zhang
\\
{\footnotesize School of Foreign Languages}
  \\
  {\footnotesize Shanghai Jiao Tong University}
  \And
  Chien-Jer Charles Lin
\\
{\footnotesize Dept.~East Asian}
\\
{\footnotesize Languages and Cultures}
  \\
  {\footnotesize Indiana University Bloomington}
  \And
  Rui Wang
  \\
  {\footnotesize Dept.~of Computer Science}
  \\
  {\footnotesize and Engineering}
  \\
  {\footnotesize Shanghai Jiao Tong University}
\\
}

\date{}

\begin{document}
\maketitle
\begin{abstract}
In this work, we revisit linguistic acceptability in the context of large language models. We introduce CoLAC - Corpus of Linguistic Acceptability in Chinese, the first large-scale  acceptability dataset for a non-Indo-European language. It is verified by native speakers and is the first acceptability dataset that comes with two sets of labels: a linguist label and a crowd label. Our experiments show that even the largest InstructGPT model performs only at chance level on CoLAC, while ChatGPT's performance (48.30 MCC) is also much below supervised models (59.03 MCC) and human (65.11 MCC). Through cross-lingual transfer experiments and fine-grained linguistic analysis, we provide detailed analysis of the model predictions and demonstrate for the first time that knowledge of linguistic acceptability can be transferred across typologically distinct languages, as well as be traced back to pre-training. Our dataset is publicly available at \url{https://github.com/huhailinguist/CoLAC}.
\end{abstract}

\section{Introduction}
The ability to distinguish acceptable sentences from unacceptable ones is central in humans' linguistic abilities, and acceptability judgements serve as the basic data in linguistic theories~\citep{1970Bever,2016Schutze}. Given an input sentence $\mathbf x$, the task is to predict whether the sentence is grammatically acceptable or not, as shown in Table~\ref{tab:example}.

Computational work on acceptability judgments is abundant in English~\citep{2018CoLA,2019BLiMP,hu-etal-2020-systematic} and other Indo-European languages~\citep{itacola,rucola}. In other languages, however, few studies have examined a computational model's ability to judge unacceptable sentences. Besides, years have passed since the publication of these works, and language models have advanced remarkably~(\citealp{2019T5,2020GPT3,2023GPT4}, \textit{inter alia}). New paradigms of learning, such as in-context learning~\citep{2020GPT3} and cross-task generalization~\citep{2021T0,2021FLAN} have risen with these new models, but acceptability judgements are yet to be systematically studied in the context of these larger models and more recent paradigms.

\begin{CJK*}{UTF8}{gkai}
\begin{table*}[t]
    \centering
    \begin{tabular}{llll}
    \toprule
        \textbf{Sentence} & \textbf{Label0} & \textbf{Label1} & \textbf{Explanation} \\
    \midrule
        一个人很聪明。 & 0 & 1 & \\
        有人难道喜欢李四吗？ & 0 & 0 & Misplacement of modal particle `难道' \\
        他唱着大声歌。 & 0 & 0 & Phrase Structure Violation \\
    \midrule
        They can sing. & 1 & - &  \\
        Maryann should leaving. & 0 & - & Morphological Violation \\
        What did Bill buy potatoes and \_? & 0 & - & Syntactic Violation \\

    \bottomrule
    \end{tabular}
    \caption{The Chinese acceptability examples are sampled from our dataset; the English sentences are from \citet{2018CoLA}. \textbf{Label0}: linguist label. \textbf{Label1}: crowd label. 0: unacceptable. 1: acceptable.}
    \label{tab:example}
\end{table*}
\end{CJK*}

In this work, we ask the following research questions:

1)
How well can neural language models perform in distinguishing grammatical acceptability in Chinese? And what is their performance with respect to different syntactic phenomena? 

2) 
Can multilingual models transfer knowledge of linguistic acceptability across typologically distinct languages? If so, where does this ability originate from? 

3) Will in-context learning be successful for acceptability judgment, compared with other language understanding tasks where it has shown remarkable success?

To answer these questions, we first introduce a large-scale dataset for acceptability in Chinese ($\S$\ref{sec:colac}), which we call Corpus of Linguistic Acceptability in Chinese (CoLAC)
. Our entire corpus of 7,495 sentences come from one syntax textbook, one linguistics handbook, and 68 journal articles, all of which have been verified by native speakers of Mandarin. Examples in the development and test sets have also been labelled with their syntactic phenomena by expert Chinese linguists, allowing for fine-grained linguistic analysis. 

We then examine the performance of monolingual and multilingual models when supervisedly fine-tuned on CoLAC training set ($\S$\ref{sec:ex-baseline}), 
and investigate the phenomenon of cross-lingual transfer in acceptability judgements ($\S$\ref{sec:ex-transfer}).
Our results show that multilingual models can benefit from supervised training in a typologically different language, even though acceptability judgement is a language-specific task that relies on the idiosyncratic syntactic phenomena of the language in question, which leads to the hypothesis that these models learn to capture linguistic acceptability during pre-training.

Finally, we experiment with few-shot learning techniques with a series of large language models from the GPT-3 family ($\S$\ref{sec:ex-icl}). Our results show that even the largest InstructGPT model only performs near chance on both CoLA and CoLAC. ChatGPT's performance, while notably better than InstructGPT, also falls well below supervised models, which are in turn below human on CoLAC.

\section{Related Work}\label{sec:related}
\paragraph{Large Language Models.} 
Pre-trained language models based on Transformer~\citep{2017Transformer} have scaled rapidly both in size and capability in recent years. Encoder models such as BERT~\citep{2018BERT}, RoBERTa~\citep{2019RoBERTa}, and its multilingual version XLM-R~\citep{2019XLM-R} have made impressive performance on NLU tasks. Encoder-decoder models such as T5~\citep{2019T5} unified understanding and generation tasks into the same text-to-text framework, and decoder-only models represented by the GPT family~\citep{2018GPT,2019GPT2,2020GPT3,2022InstructGPT,2023GPT4} have scaled to hundreds of billions of parameters, unlocking emergent abilities~\citep{2022emergent} such as in-context learning.

\paragraph{Few-shot Learning.} 
\citet{2019GPT2} found that language models implicitly learn to multitask during pre-training by directly applying the pre-trained model to downstream tasks. Instruction tuning~\citep{2021T0,2021FLAN,2022FLAN} exploits this ability of language models and explicitly train them to activate zero-shot generalization ability on unseen tasks. In-context learning~\citep{2020GPT3}, on the other hand, provide the models with input-output examples of a new task in the models' input context to help them quickly adapt to the task.

\paragraph{Linguistic Acceptability.} 
To the best of our knowledge, most if not all corpora 
on acceptability judgments collected from syntax textbooks or journal articles are built for Indo-European languages, namely CoLA for English~\citep{2018CoLA}, ItaCoLA for Italian~\citep{itacola}, and RuCoLA for Russian~\citep{rucola}.
There has been other works, including BLiMP~\citep{2019BLiMP}, CLiMP~\citep{2021CLiMP} and SyntaxGym~\citep{hu-etal-2020-systematic}, that address acceptability tests in language models from the aspect of linguistic minimal pairs, either semi-automatically constructed or manually created.  


CoLA was included in the GLUE benchmark~\citep{2018GLUE}, but excluded from SuperGLUE~\citep{2019SGLUE} based on the rationale that language models' performance on this task had surpassed human at that time~\citep{2019XLNet}, even though this performance was far from perfect, and larger models later reported high inter-run variance on it~\citep{2019T5}. As a result, this particular linguistic aspect of language models has remained understudied in recent years. To the best of our knowledge, most work of LLMs did not report experimental results on this task~\citep{2020GPT3,2021Gopher,2022MT-NLG,2022LaMDA,2022Chinchilla,2022PaLM,2022OPT,2022BLOOM}.
A notable exception is 
\citet{sinha-etal-2023-language}, which investigates how different contexts (context length, grammaticality type, etc.) for test examples affect the performance of language models (GPT-2 and five variants of OPT) in the minimal pair paradigm.

\section{CoLAC: Corpus of Linguistc Acceptability in Chinese}\label{sec:colac}

We present Corpus of Linguistic Acceptability in Chinese (CoLAC), the first large-scale  acceptability dataset in a non-Indo-European language handcrafted by linguists to evaluate the grammatical proficiency of language models. Our dataset consists of 7,495 sentences collected from one syntax textbook, one linguistics handbook, and 68 linguistics journal articles, all verified by native speakers of Mandarin. 




\subsection{Data Collection}\label{sec:colac-collection}
We compiled CoLAC from three sources: 

1) \textbf{Textbook}: We gathered more than one thousand examples from \textit{The Syntax of Chinese}~\citep{2009chinesesyntax}, collected by \citet{2020chinesesyntax}, who partitioned the sentences into two categories: minimal pairs and stand-alone sentences. \citet{2020chinesesyntax} verified all minimal pairs with native speakers using a 7-point Likert Scale.

2) \textbf{Journal}: \citet{acceptability-hu} collected about seven thousand examples from 68 articles in linguistic journals, and verified 674 sentences (337 minimal pairs) with native speakers. We verify the rest examples.

3) \textbf{Handbook}: We also collected 1,029 sentences from 10 chapters in \textit{Handbook of Chinese Linguistics}~\citep{2018handbook}, which focuses on specific syntactic phenomena.

Following \citet{2018CoLA}, we manually checked every sentence in the collected data, and removed the examples that fall into one of the following categories: 
1) the example is about semantic or pragmatic anomaly rather than syntactic anomaly; 
2) the ungrammaticality is associated with a specific prosodic or stress pattern; 
3) the sentence is from classical Chinese or a dialect; 
4) the example involves a conversation; 
5) the grammaticality of the sentence relies upon co-reference or co-indexing of certain lexical items. 

\subsection{Verification}
For all sentences that have not been verified (i.e. all sentences except for those from \citet{2009chinesesyntax}, we ask at least five native Mandarin speakers to rate their naturalness on a 4-point Likert Scale, with one being very unnatural, two being unnatural, three being natural and four being very natural. In total, 485 participants contributed to the annotation process.



After verification from crowd workers, every example sentence has two labels:

1) a single label from the linguist who proposed the example\footnote{Note that this label is not from a single linguist, as we collected examples from one syntax textbook, one handbook for Chinese syntax and about 70 journal articles authored by different theoretical syntacticians. }, which we call \textbf{linguist label} (label0),

2)  a \textbf{crowd label} (label1), mapped from the mean ratings from other native speakers of Mandarin Chinese. 

\begin{table}[]
    \centering
    \begin{tabular}{m{0.3cm}ccccc}\toprule
         & split & size & accept. \%  & mean len \\\midrule
        \multirow{3}{0.3cm}{\rotatebox[origin=c]{90}{CoLAC}} & train  & 6072 & 66.9 & 11.8   \\ 
         & dev  & 492 & 63.4 & 11.1 \\
         & test  & 931 & 64.1 & 11.2 \\\midrule
        \multirow{3}{0.3cm}{\rotatebox[origin=c]{90}{CoLA}} & train  & 8551 & 70.4 & 34.0 \\
         & dev  & 527 & 69.3 & 33.3 \\
         & test & 516 & 68.6 & 36.6 \\\bottomrule
    \end{tabular}
    \caption{Statistics of CoLAC (label1), compared with CoLA~\citep{2018CoLA}: train/development/test split, percentage of acceptable sentences, and mean length of the sentences in characters.}
    \label{tab:colac:distribution}
\end{table}

Unlike \citet{2018CoLA}, who used the linguist label as the ground truth of their dataset, we adopt the crowd label as our ground truth label. The rationale is that we believe the linguist label reflects ``grammaticality'', which is usually deemed to reflect language competence that generative syntacticians are interested in. On the other hand, linguistic ``acceptability'' is more often associated with language performance and language use, and therefore better reflected by native speakers' acceptability judgments, that is our crowd label. 
This discrepancy is especially notable in the case of colloquial languages such as ``\begin{CJK*}{UTF8}{gkai}\texttt{张三的脸稍微红红的}\end{CJK*}''\footnote{which literally translates to ``Zhangsan's face is a bit red.''}, which is deemed to be grammatically correct by \citet{2013redred}, but is unanimously rated unacceptable by six of our participants.
Furthermore, adopting crowd ratings as the ground truth label for CoLAC also enables us to directly and intrinsically estimate human performance on CoLAC from the subjects that contributed to the labels, as we elaborate in detail in $\S$\ref{sec:colac-human}.

The final dataset that we collected contains 7,495 sentences, all verified by native speakers. As shown in Table \ref{tab:colac:distribution}, we split CoLAC into training, development, and test sets with the same label ditributions.

\subsection{Label Descriptions}\label{sec:colac-label}

The syntactic phenomena covered in CoLAC are classified into 18 categories. For each category, we provide a short explanation and a pair of illustrative examples in Table \ref{tab:labels}.

\subsection{Human Performance}\label{sec:colac-human}

Following the English CoLA~\citep{2018CoLA}, we use Matthews Correlation Coefficient (MCC) as the measure for both human and model performance, as the two classes are not balanced. MCC ranges from -1 (perfect negative correlation) to 0 (no correlation) to 1 (perfect positive correlation); however, in this work we time it by 100 and present a number between -100 to 100.

We measure MCC of each subject's annotation for his or her assigned sentence list (comprising about 120 sentences) against the final aggregated labels, and then take the average of all 485 contributors' MCC.

The average MCC computed this way is 65.11.
In comparison, human MCC reported by \citet{2018CoLA} is 69.7. This suggests that linguistic acceptability is indeed an difficult task subject to individual interpretation. Here we additionally note that while many textual tasks - sentiment analysis, summarization, translation, etc. - are also subjective to some extent, linguistic acceptability is particularly tricky due to the involvement of more specialized grammatical inflections and 
can be affected by many factors, e.g., grammaticality, processing difficulty, dialectal background among others \citep{yao2018}.




\section{Experimental setup}
\subsection{Supervised Baselines}\label{sec:ex-baseline}
As a supervised baseline, we fine-tune XLM-R~\citep{2019XLM-R} and Chinese RoBERTa~\citep{2020ChineseRoBERTa} on CoLAC training set. Acknowledging that encoder-decoder and decoder-only models have taken the lead in the recent development of language models, both of which convert classification tasks into a text-to-text format, we also consider prompt-tuning XLM-R using the following template, where the model is trained to output \texttt{Yes} or \texttt{No} at the masked position:
\begin{lstlisting}[breaklines]
{sent} </s> Is this sentence acceptable or not? <mask>.
\end{lstlisting}

To avoid overfitting on the test set, we report results on CoLAC development set throughout the paper, unless otherwise stated. To mitigate the effect of inter-run variance, we report the median of five random seeds. Further training details are given in appendix~\ref{sec:training-detail}.

\subsection{Transfer between CoLA and CoLAC}\label{sec:ex-transfer}
With the introduction of the first large-scale non-Indo-European acceptability dataset, we are now able to better study the effects of cross-lingual transfer in linguistic acceptability. Recent works in multi-task training have found that with a diverse enough set of training tasks, the number of each task's training samples does not need to be very large, and its contribution to the model's performance saturates at about 64~\citep{2022SuperNatural}, which suggests that multi-task training only serves as a conduit to activate langauge models' knowledge learned during pre-training~\citep{2019GPT2} instead of acquisition of new knowledge. Observing that CoLA and CoLAC are essentially two different tasks\footnote{Most grammatical phenomena covered in CoLA and CoLAC are specific to English and Chinese and do not overlap with each other. See $\S$\ref{sec:analysis} for more examples.}, we train XLM-R on downsampled CoLAC and CoLA training sets to observe the emergence of the ability to perform acceptability judgements in the other langauge.

\subsection{Few-shot Learning}\label{sec:ex-icl}
We experiment with OpenAI's GPT series to provide a few-shot learning baseline on CoLAC. We also repeat the same experiments on CoLA to facilitate future studies in this area. Prompt details are given in Appendix~\ref{sec:prompt}.
\section{Results}
\subsection{Baselines}
\begin{table}[th]
    \centering
    \begin{tabular}{ccc}
    \toprule
        \textbf{Model} & Dev MCC & Test MCC \\
    \midrule
        XLM-R & 56.45 & 54.08\\
        XLM-R$^p$ & 54.66 & - \\
        Zh RoBERTa & 56.95 & -\\
        Zh RoBERTa$^e$ & 58.07 & 59.03\\
        ChatGPT$^5$ & 47.82 & 48.30\\
    \bottomrule
    \end{tabular}
    \caption{Baseline performance on CoLAC development and test sets. Results on test set are reported sparingly. $^p$: prompt-tuned. $^e$: ensembled. $^5$: 5-shot.}
    \label{tab:baseline}
\end{table}

In Table~\ref{tab:baseline}, we present the results of supervised models. Prompt-tuned XLM-R performs comparatively with fine-tuning, and Chinese RoBERTa performs marginally better. In an effort to elicit the best possible performance from these models without loss of generality, we also ensemble the predictions of five monolingual models, which scores 58.07 on development set and 59.03 on test set.

\subsection{Few-shot Learning of LLM}
\begin{table}[t]
    \centering
    \small
    \scalebox{.8}{
    \begin{tabular}{c|ccc|ccc}
    \toprule
        & \multicolumn{3}{c}{CoLAC} & \multicolumn{3}{c}{CoLA}\\
        \textbf{Model} & 0-shot & 1-shot & 5-shot & 0-shot & 1-shot & 5-shot  \\
    \midrule
        350M & 0.00 & 0.00  & 0.00 & 0.00 & 0.00  & 0.00  \\
        1.3B & 0.00 & 0.00 & 1.38 & 0.00& 0.00 &  6.54 \\
        6.7B & -4.85 & 4.54 & 0.23 &-2.90 & -5.61 & -5.72 \\    
        175B & 0.00 & 2.52 & 2.52 & 0.00 & 0.00 & 0.00\\
        ChatGPT & -2.59 & 42.81 & 47.82 &  -3.37 & 54.69 & 64.56  \\
    \bottomrule
    \end{tabular}
    }
    \caption{Performance (MCC) of InstructGPT and ChatGPT on CoLAC / CoLA development set. 0 entails random guessing.}
    \label{tab:gpt-colac}
\end{table}


In Table~\ref{tab:gpt-colac}, we present the performance of four InstructGPT models as well as ChatGPT on CoLAC development set. Surprisingly, even though InstructGPT and its predecessor GPT-3 have demonstrated strong in-context learning performance on a wide range of benchmarks~\citep{2020GPT3,2022InstructGPT}, they fail to exhibit any informed judgement on linguistic acceptability. ChatGPT is the only exception, which scores -2.59 in the zero-shot setting, but quickly improves to 42.18 when provided with an example to help it understand the task. Even with 5 in-context examples, though, its performance is still below supervised models, as shown in Table~\ref{tab:baseline}.

\begin{figure}
    \centering
    \includegraphics[width=1\linewidth]{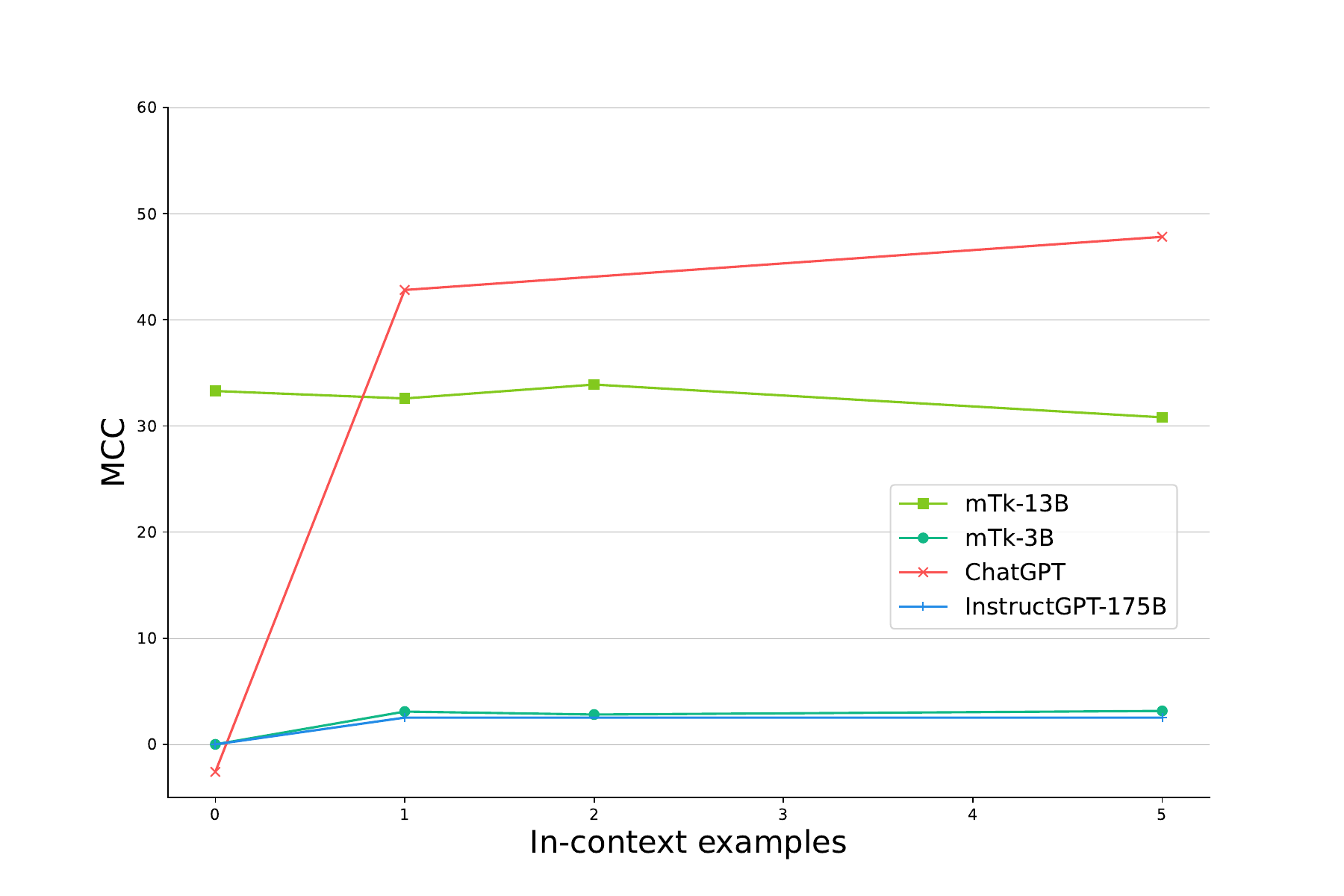}
    \caption{In-context learning performance on CoLAC development set.}
    \label{fig:icl}
\end{figure}

\begin{figure*}[t]
    \centering
    \includegraphics[width=1\textwidth]{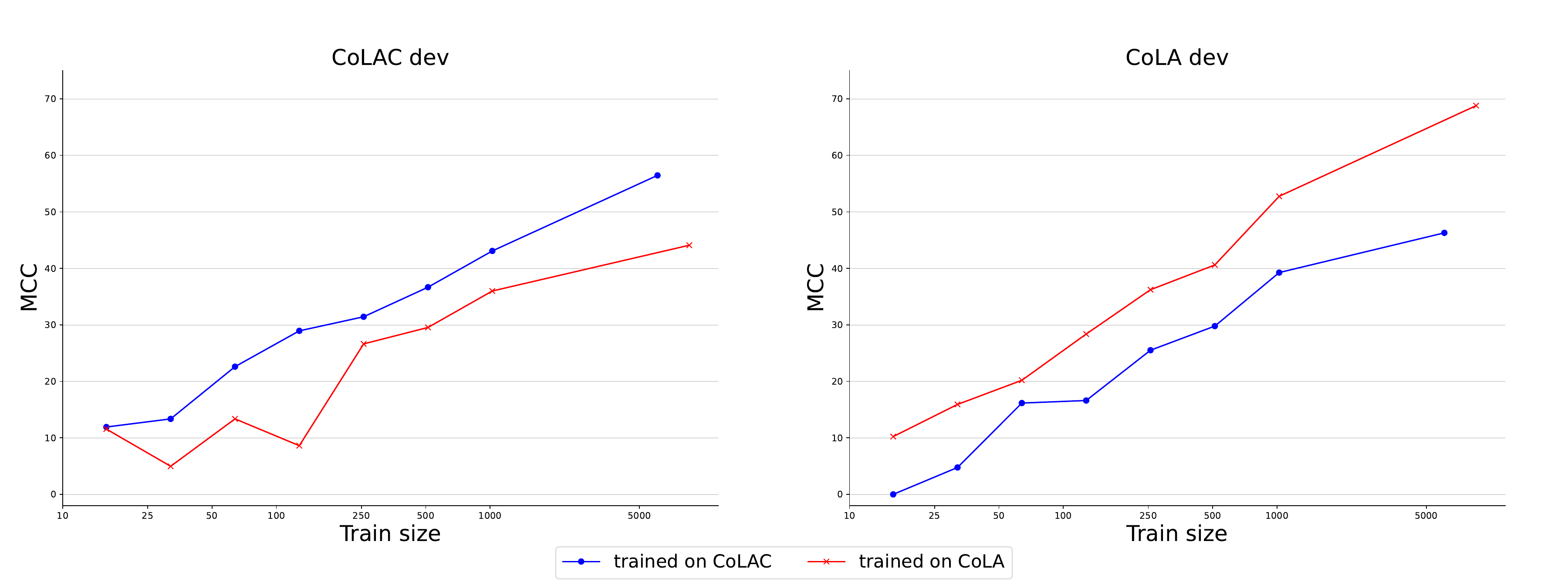}
    \caption{In-domain and out-of-domain performance of XLM-R trained on CoLA and CoLAC, plotted against log-scale training set size. Each language is considered to be an individual domain (a different task).}
    \label{fig:transfer}
\end{figure*}

A more surprising finding is that experiments on English CoLA yield the same results, as exhibited in Table~\ref{tab:gpt-colac}. This gives us the caution that the generalization capability of LLMs has probably been overestimated in many aspects, especially in those involving specialized knowledge - such as linguistic acceptability - that does not appear frequently in instruction-tuning datasets. It also reminds us that sometimes insights of next-generation LLMs can actually be gained from revisiting classical benchmarks.\footnote{As we noted in $\S$\ref{sec:related}, no work of LLM since GPT-3 has actually bothered with the original tasks in GLUE.}

For completeness, we also report the results of mTk-Instruct~\citep{2022SuperNatural}, a variant of mT5~\citep{2020mT5} that has been explicitly trained for linguistic acceptability on CoLA (see Appendix~\ref{sec:prompt} for details). The results are plotted in Figure~\ref{fig:icl}. In the zero-shot setting, mTk-Instruct exhibits non-trivial performance, but unlike ChatGPT it does not benifit from more in-context examples, probably due to the rigid instruction template of its training data.

With mTk-Instruct, we conducted ablation studies concerning the selection and ordering of in-context examples, and find that they have limited impact on the outcome.

\subsection{Tracing the Origin of Acceptability Judgement Ability}

In Figure~\ref{fig:transfer}, we plot the results of in-domain and out-of-domain\footnote{Here we consider English and Chinese as two different domains; i.e. we consider CoLA and CoLAC to be two related but different tasks.} performance of XLM-R trained on CoLA and CoLAC training sets that are downsampled to varying degrees, but all with the same amount of compute (see appendix~\ref{sec:training-detail} for details).

From the figure, we observe that in-domain performance demonstrates a smooth log-linear curve with respect to the amount of training data, conforming to the scaling law~\citep{2020scaling,2022Chinchilla}. Out-of-domain performance, however, does not exhibit such clear patterns. It manifests a phase change at 256 training samples when transferring from English to Chinese, similar to the behavior of emergent abilities~\citep{2022emergent}, but not when transferring from Chinese to English. Also, the increase of out-of-domain performance on both CoLA and CoLAC significantly slows down beyond 1024 samples, while in-domain performance maintains a steady growth. These behaviors are similar in essence to multi-task fine-tuning~\citep{2021T0,2021FLAN,2022SuperNatural}, studies of which have found that zero-shot generalization performance on unseen tasks saturate at only dozens of examples per training task.

Based on these observations and the fact that the underlying syntax of English and Chinese are very different and thus unlikely to be directly transferred during fine-tuning, we hypothesize that language models must have acquired some aspects of linguistic acceptability during pre-training, even for small models such as XLM-R (which has 550M parameters). However, with only two datasets, we are not able to ascertain how much of this ability is learned during pre-training and how much comes from cross-lingual transfer during fine-tuning.  
We leave a more systematic analysis on a more diverse set of languages to future work.
\section{Analysis}\label{sec:analysis}

\begin{figure*}[th]
    \centering
    \scalebox{.8}{
    \includegraphics[width=1\textwidth]{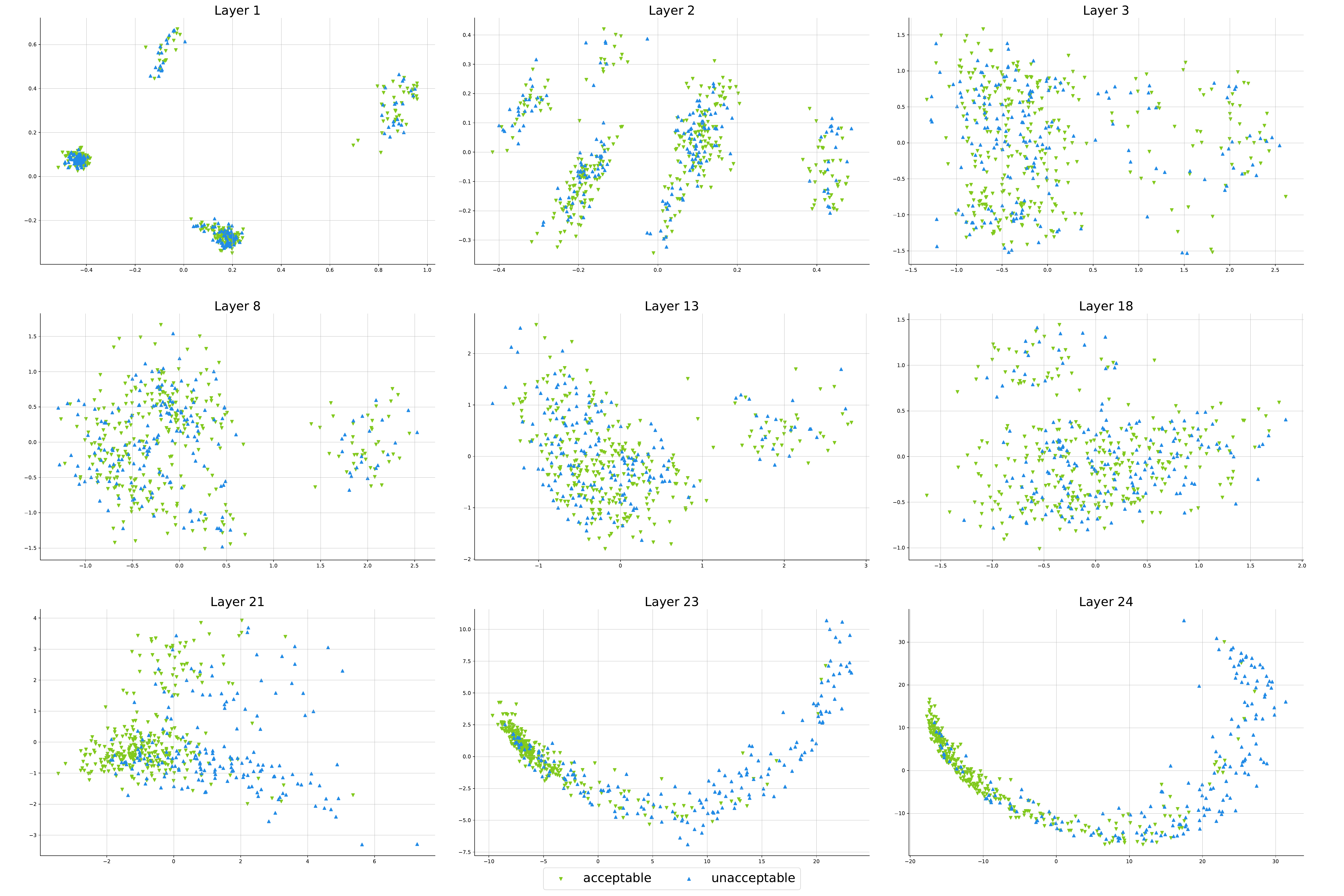}
    }
    \caption{Intermediate representations of CoLAC development set in XLM-R.}
    \label{fig:inner}
\end{figure*}

\begin{figure}[th]
    \centering
    \includegraphics[width=1\linewidth]{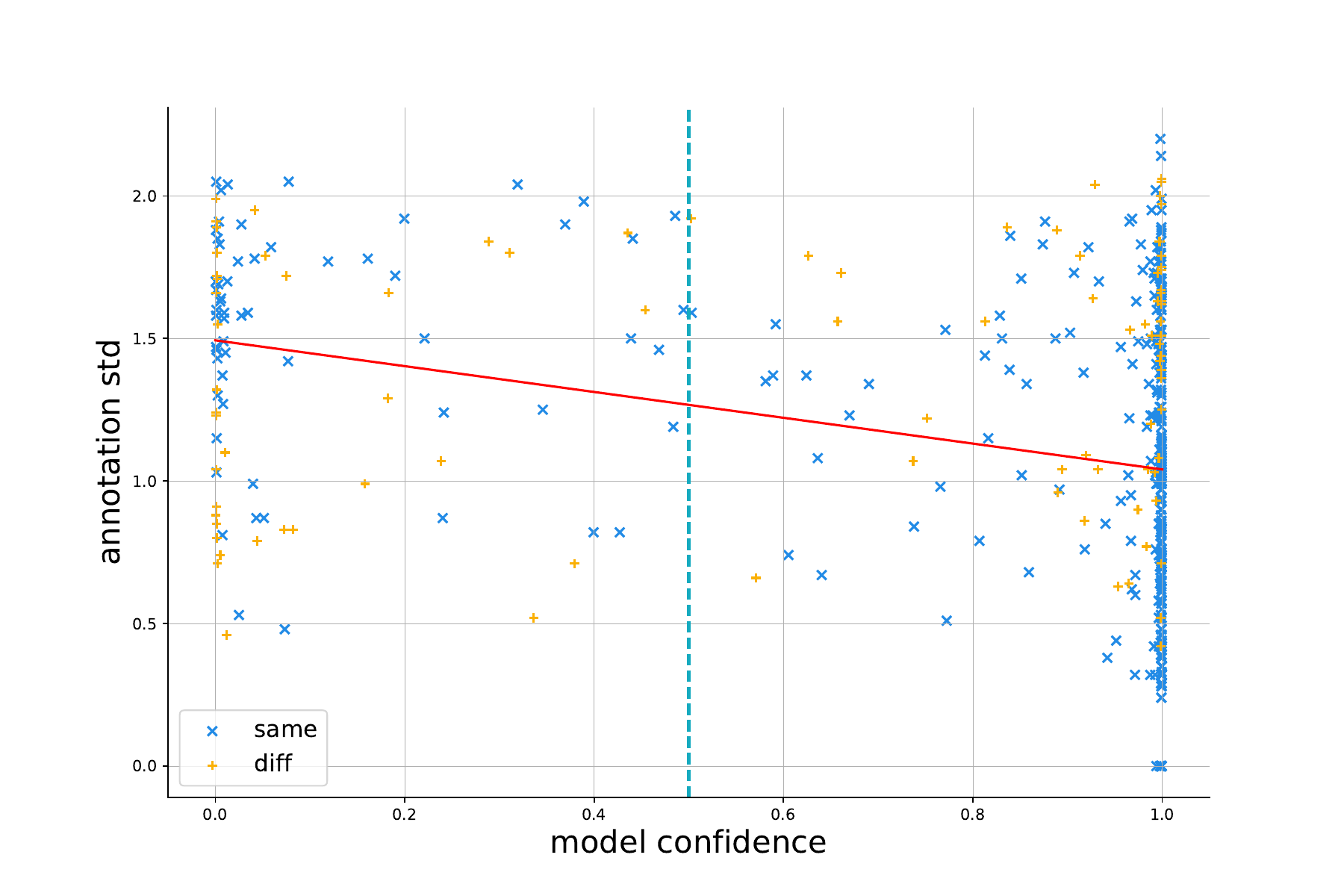}
    \caption{Relationship between std.~of crowd annotation and XLM-R's confidence for the correct label1 on CoLAC development set. Points on the right side are classified correctly by the model, and those on the left side are not. Samples assigned with the same/different label0 and label1 are indicated by \texttt{same/diff}.}
    \label{fig:agree}
\end{figure}

\begin{figure*}
    \centering
    \includegraphics[width=1\textwidth]{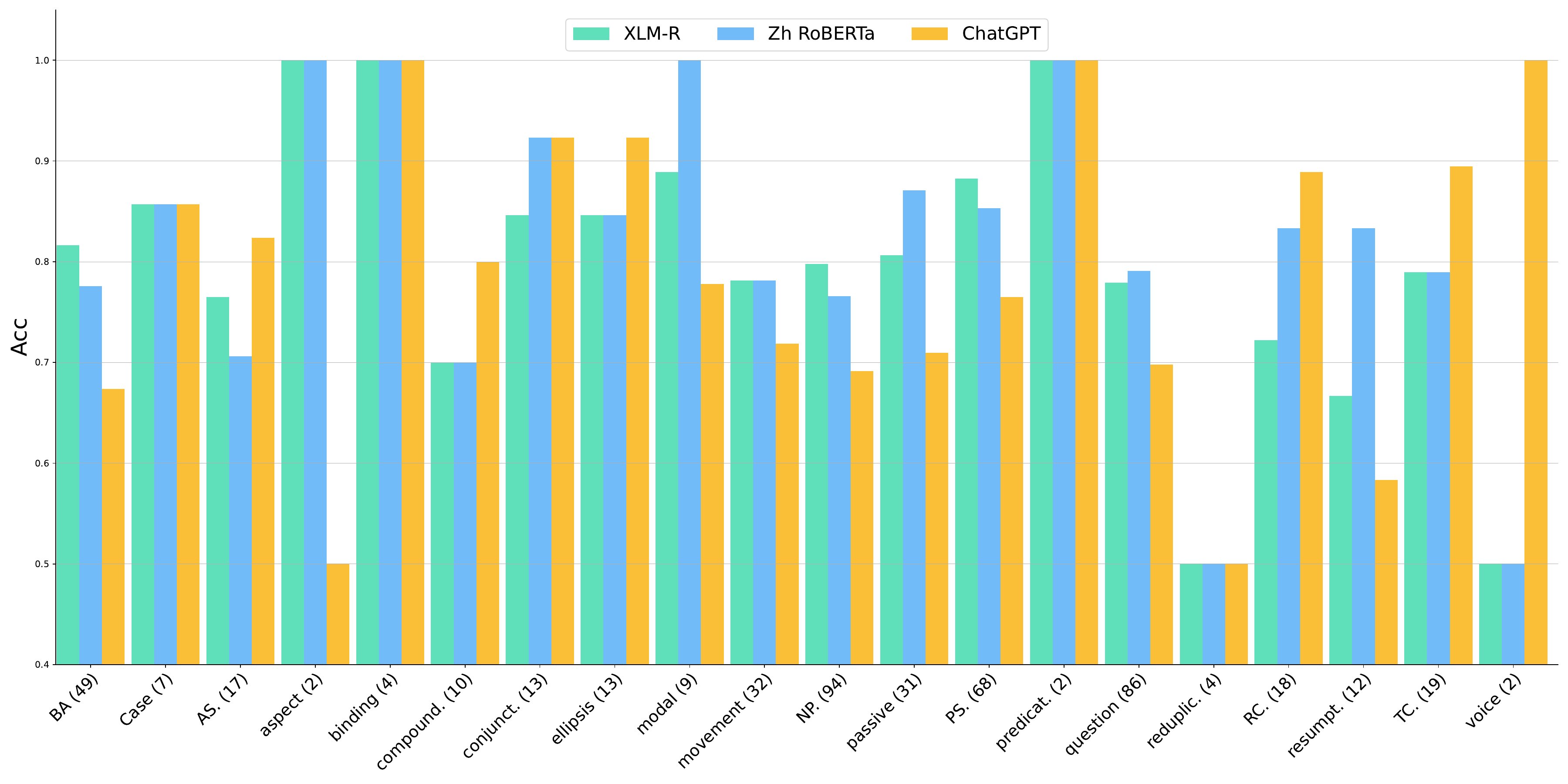}
    \caption{Performance of XLM-R, Chinese RoBERTa, and ChatGPT on CoLAC development set, categorized into the syntactic phenomena in Table~\ref{tab:labels}. The number after each syntactic phenomenon indicates the number of corresponding samples in the data.}
    \label{fig:syntax}
\end{figure*}

\subsection{Intermediate Representations}

In Figure~\ref{fig:inner}, we visualize the intermediate representations of CoLAC development set in a fine-tuned XLM-R. We observe that in the first two layers, the sentences are grouped into four distinct clusters.
Three of these four clusters represent common syntactic structures in Chinese: most sentences in the left cluster are \textit{inverted}, where the object or adjective phrases are moved to the beginning of the sentences, while sentences in the top cluster are all \textit{interrogatives}. Sentences in the right cluster are interrogatives as well, but with an emphasis on \textit{disjunctive question}. As the sentences are propagated further up the network, these generic syntactical phenomena are gradually transformed into acceptability-specific features, as Figure~\ref{fig:inner} demonstrates.

An analogous analysis on CoLA development set also reveals a similar observation, where the sentences are grouped into two distinct clusters in the intermediate layers, with one cluster accommodating interrogatives and exclamations, and the other cluster containing declarative sentences. These generic features are also transformed into features that can distinguish acceptability in later layers. These findings are complementary with previous findings that Transformers learn to capture local syntax in the lower layers, and to transform it into more complex semantics in higher layers~\citep{2019pipeline}.


\subsection{Annotator Disagreement and Model Performance}

During the annotation process, we found that some samples in CoLAC received low inter-annotator agreement. In Figure~\ref{fig:agree}, we plot the annotation disagreement against XLM-R's prediction confidence, and also highlight the samples that received distinctive linguist and crowd labels. 
From the figure, we observe that inter-annotator disagreement is only weakly correlated with model confidence, with a least squares slope of -0.45. Thus, we decide to keep these samples in the dataset to simulate real-world scenarios, where there actually exists many borderline sentences that are hard to classify.

From Figure~\ref{fig:agree}, we also observe that samples with different crowd label and linguist label (marked as \texttt{diff}) take up a larger portion of wrongly classified samples (0.37) compared to correctly classified samples (0.15). 
This suggests that when the crowd and linguist labels of an example differ, models are more prone to wrongly classify it.  
This corroborates with previous findings that ambiguous samples are harder to learn~\citep{swayamdipta-etal-2020-dataset-cartography} and highlights the need for a more careful annotation scheme for NLP datasets, especially for sentence classification task, as suggested by some recent work in natural language inference, where both a strict logic label and a loose commonsense label are provided~\citep{kalouli2023}, or modify a two/three-way classification task to a more fine-grained regression task~\citep{unli}. 

\subsection{Syntactic Categories}

In Figure~\ref{fig:syntax}, we categorize the sentences in CoLAC development set into twenty syntactic categories, and plot the performance of three models on each category. Among these models, ChatGPT excels in \textsc{argument structure}, \textsc{compounding}, \textsc{ellipsis}, \textsc{relative clause}, and \textsc{topic construction}, which suggests that these structures probably appear frequently in dialogues, for which ChatGPT is optimized. The monolingual model, on the other hand, is notably better at \textsc{modal}, \textsc{passive}, \textsc{question}, and \textsc{resumptive}, which may provide insights about the syntactic structures that future multilingual models should pay more attention to when adapted for Chinese applications.

To our surprise, on the \textsc{question} category, ChatGPT's performance is significantly lower than the other two small models. In an effort to understand this behavior, we manually checked the sentences that ChatGPT classified incorrectly, and here are two typical examples\footnote{In our experiments, ChatGPT sometimes provides an additional explanation apart from the answer, but not for any sentence in this category.}:

\begin{CJK*}{UTF8}{gkai}
你到底想看电影还是打麻将吗？

有人难道喜欢李四吗？
\end{CJK*}

Both of these two sentences are clearly unacceptable to native Chinese speakers, because of the semantic incompatibility of the polar interrogative particle ``\begin{CJK*}{UTF8}{gkai}吗\end{CJK*}'' with disjunctive questions, and the misplacement of the Chinese modal particle ``\begin{CJK*}{UTF8}{gkai}难道\end{CJK*}''. We hypothesize that ChatGPT may have learned to ignore such errors as it has been optimized for real-world dialogue applications.

\section{Conclusion}
We presented CoLAC, the first large-scale non-Indo-European acceptability dataset that comes with two sets of labels. 
Experiments show that CoLAC is a quite challenging dataset, with ChatGPT performing well below supervised models or human, but benefiting from in-context examples. Through cross-lingual transfer experiments, we demonstrated for the first time that concepts of linguistic acceptability can be transferred across typologically different languages, and may be acquired during pre-training. In the future, we intend to extend the existing linguistic acceptability datasets into a multilingual benchmark containing more languages.

\bibliography{tacl2021}

\begin{thebibliography}{48}
\expandafter\ifx\csname natexlab\endcsname\relax\def\natexlab#1{#1}\fi

\bibitem[{Bever(1970)}]{1970Bever}
Thomas Bever. 1970.
\newblock The influence of speech performance on linguistic structures.
\newblock \emph{G. B. Flores d'Arcais, W. J. M. Levelt, Hrsg., Advances in
  psydiolinguistics}, pages 4--30.

\bibitem[{Brown et~al.(2020)Brown, Mann, Ryder, Subbiah, Kaplan, Dhariwal,
  Neelakantan, Shyam, Sastry, Askell, Agarwal, Herbert{-}Voss, Krueger,
  Henighan, Child, Ramesh, Ziegler, Wu, Winter, Hesse, Chen, Sigler, Litwin,
  Gray, Chess, Clark, Berner, McCandlish, Radford, Sutskever, and
  Amodei}]{2020GPT3}
Tom~B. Brown, Benjamin Mann, Nick Ryder, Melanie Subbiah, Jared Kaplan,
  Prafulla Dhariwal, Arvind Neelakantan, Pranav Shyam, Girish Sastry, Amanda
  Askell, Sandhini Agarwal, Ariel Herbert{-}Voss, Gretchen Krueger, Tom
  Henighan, Rewon Child, Aditya Ramesh, Daniel~M. Ziegler, Jeffrey Wu, Clemens
  Winter, Christopher Hesse, Mark Chen, Eric Sigler, Mateusz Litwin, Scott
  Gray, Benjamin Chess, Jack Clark, Christopher Berner, Sam McCandlish, Alec
  Radford, Ilya Sutskever, and Dario Amodei. 2020.
\newblock \href
  {https://proceedings.neurips.cc/paper/2020/hash/1457c0d6bfcb4967418bfb8ac142f64a-Abstract.html}
  {Language models are few-shot learners}.
\newblock In \emph{Advances in Neural Information Processing Systems 33: Annual
  Conference on Neural Information Processing Systems 2020, NeurIPS 2020,
  December 6-12, 2020, virtual}.

\bibitem[{Chen et~al.(2020{\natexlab{a}})Chen, Jiang, Poliak, Sakaguchi, and
  Van~Durme}]{unli}
Tongfei Chen, Zheng~Ping Jiang, Adam Poliak, Keisuke Sakaguchi, and Benjamin
  Van~Durme. 2020{\natexlab{a}}.
\newblock Uncertain natural language inference.
\newblock In \emph{Proceedings of the 58th Annual Meeting of the Association
  for Computational Linguistics}, pages 8772--8779.

\bibitem[{Chen et~al.(2020{\natexlab{b}})Chen, Xu, and Xie}]{2020chinesesyntax}
Zhong Chen, Yuhang Xu, and Zhiguo Xie. 2020{\natexlab{b}}.
\newblock Assessing introspective linguistic judgments quantitatively: the case
  of the syntax of chinese.
\newblock \emph{Journal of East Asian Linguistics}, 29(3):311--336.

\bibitem[{Chowdhery et~al.(2022)Chowdhery, Narang, Devlin, Bosma, Mishra,
  Roberts, Barham, Chung, Sutton, Gehrmann, Schuh, Shi, Tsvyashchenko, Maynez,
  Rao, Barnes, Tay, Shazeer, Prabhakaran, Reif, Du, Hutchinson, Pope, Bradbury,
  Austin, Isard, Gur{-}Ari, Yin, Duke, Levskaya, Ghemawat, Dev, Michalewski,
  Garcia, Misra, Robinson, Fedus, Zhou, Ippolito, Luan, Lim, Zoph, Spiridonov,
  Sepassi, Dohan, Agrawal, Omernick, Dai, Pillai, Pellat, Lewkowycz, Moreira,
  Child, Polozov, Lee, Zhou, Wang, Saeta, Diaz, Firat, Catasta, Wei,
  Meier{-}Hellstern, Eck, Dean, Petrov, and Fiedel}]{2022PaLM}
Aakanksha Chowdhery, Sharan Narang, Jacob Devlin, Maarten Bosma, Gaurav Mishra,
  Adam Roberts, Paul Barham, Hyung~Won Chung, Charles Sutton, Sebastian
  Gehrmann, Parker Schuh, Kensen Shi, Sasha Tsvyashchenko, Joshua Maynez,
  Abhishek Rao, Parker Barnes, Yi~Tay, Noam Shazeer, Vinodkumar Prabhakaran,
  Emily Reif, Nan Du, Ben Hutchinson, Reiner Pope, James Bradbury, Jacob
  Austin, Michael Isard, Guy Gur{-}Ari, Pengcheng Yin, Toju Duke, Anselm
  Levskaya, Sanjay Ghemawat, Sunipa Dev, Henryk Michalewski, Xavier Garcia,
  Vedant Misra, Kevin Robinson, Liam Fedus, Denny Zhou, Daphne Ippolito, David
  Luan, Hyeontaek Lim, Barret Zoph, Alexander Spiridonov, Ryan Sepassi, David
  Dohan, Shivani Agrawal, Mark Omernick, Andrew~M. Dai,
  Thanumalayan~Sankaranarayana Pillai, Marie Pellat, Aitor Lewkowycz, Erica
  Moreira, Rewon Child, Oleksandr Polozov, Katherine Lee, Zongwei Zhou, Xuezhi
  Wang, Brennan Saeta, Mark Diaz, Orhan Firat, Michele Catasta, Jason Wei,
  Kathy Meier{-}Hellstern, Douglas Eck, Jeff Dean, Slav Petrov, and Noah
  Fiedel. 2022.
\newblock \href {https://doi.org/10.48550/arXiv.2204.02311} {Palm: Scaling
  language modeling with pathways}.
\newblock \emph{CoRR}, abs/2204.02311.

\bibitem[{Chung et~al.(2022)Chung, Hou, Longpre, Zoph, Tay, Fedus, Li, Wang,
  Dehghani, Brahma, Webson, Gu, Dai, Suzgun, Chen, Chowdhery, Narang, Mishra,
  Yu, Zhao, Huang, Dai, Yu, Petrov, Chi, Dean, Devlin, Roberts, Zhou, Le, and
  Wei}]{2022FLAN}
Hyung~Won Chung, Le~Hou, Shayne Longpre, Barret Zoph, Yi~Tay, William Fedus,
  Eric Li, Xuezhi Wang, Mostafa Dehghani, Siddhartha Brahma, Albert Webson,
  Shixiang~Shane Gu, Zhuyun Dai, Mirac Suzgun, Xinyun Chen, Aakanksha
  Chowdhery, Sharan Narang, Gaurav Mishra, Adams Yu, Vincent~Y. Zhao, Yanping
  Huang, Andrew~M. Dai, Hongkun Yu, Slav Petrov, Ed~H. Chi, Jeff Dean, Jacob
  Devlin, Adam Roberts, Denny Zhou, Quoc~V. Le, and Jason Wei. 2022.
\newblock \href {https://doi.org/10.48550/arXiv.2210.11416} {Scaling
  instruction-finetuned language models}.
\newblock \emph{CoRR}, abs/2210.11416.

\bibitem[{Conneau et~al.(2020)Conneau, Khandelwal, Goyal, Chaudhary, Wenzek,
  Guzm{\'{a}}n, Grave, Ott, Zettlemoyer, and Stoyanov}]{2019XLM-R}
Alexis Conneau, Kartikay Khandelwal, Naman Goyal, Vishrav Chaudhary, Guillaume
  Wenzek, Francisco Guzm{\'{a}}n, Edouard Grave, Myle Ott, Luke Zettlemoyer,
  and Veselin Stoyanov. 2020.
\newblock \href {https://doi.org/10.18653/v1/2020.acl-main.747} {Unsupervised
  cross-lingual representation learning at scale}.
\newblock In \emph{Proceedings of the 58th Annual Meeting of the Association
  for Computational Linguistics, {ACL} 2020, Online, July 5-10, 2020}, pages
  8440--8451. Association for Computational Linguistics.

\bibitem[{Cui et~al.(2020)Cui, Che, Liu, Qin, Wang, and
  Hu}]{2020ChineseRoBERTa}
Yiming Cui, Wanxiang Che, Ting Liu, Bing Qin, Shijin Wang, and Guoping Hu.
  2020.
\newblock \href {https://doi.org/10.18653/v1/2020.findings-emnlp.58}
  {Revisiting pre-trained models for chinese natural language processing}.
\newblock In \emph{Findings of the Association for Computational Linguistics:
  {EMNLP} 2020, Online Event, 16-20 November 2020}, volume {EMNLP} 2020 of
  \emph{Findings of {ACL}}, pages 657--668. Association for Computational
  Linguistics.

\bibitem[{Devlin et~al.(2019)Devlin, Chang, Lee, and Toutanova}]{2018BERT}
Jacob Devlin, Ming{-}Wei Chang, Kenton Lee, and Kristina Toutanova. 2019.
\newblock \href {https://doi.org/10.18653/v1/n19-1423} {{BERT:} pre-training of
  deep bidirectional transformers for language understanding}.
\newblock In \emph{Proceedings of the 2019 Conference of the North American
  Chapter of the Association for Computational Linguistics: Human Language
  Technologies, {NAACL-HLT} 2019, Minneapolis, MN, USA, June 2-7, 2019, Volume
  1 (Long and Short Papers)}, pages 4171--4186. Association for Computational
  Linguistics.

\bibitem[{Hoffmann et~al.(2022)Hoffmann, Borgeaud, Mensch, Buchatskaya, Cai,
  Rutherford, de~Las~Casas, Hendricks, Welbl, Clark, Hennigan, Noland,
  Millican, van~den Driessche, Damoc, Guy, Osindero, Simonyan, Elsen, Rae,
  Vinyals, and Sifre}]{2022Chinchilla}
Jordan Hoffmann, Sebastian Borgeaud, Arthur Mensch, Elena Buchatskaya, Trevor
  Cai, Eliza Rutherford, Diego de~Las~Casas, Lisa~Anne Hendricks, Johannes
  Welbl, Aidan Clark, Tom Hennigan, Eric Noland, Katie Millican, George van~den
  Driessche, Bogdan Damoc, Aurelia Guy, Simon Osindero, Karen Simonyan, Erich
  Elsen, Jack~W. Rae, Oriol Vinyals, and Laurent Sifre. 2022.
\newblock \href {https://doi.org/10.48550/arXiv.2203.15556} {Training
  compute-optimal large language models}.
\newblock \emph{CoRR}, abs/2203.15556.

\bibitem[{Hu et~al.(in preparation)Hu, Li, Ma, Huang, and
  Lin}]{acceptability-hu}
Hai Hu, Aini Li, Yini Ma, Jiahui Huang, and Chien-Jer~Charles Lin. in
  preparation.
\newblock Grammaticality in {Chinese}: Dialectal influences and sources of
  variability.

\bibitem[{Hu et~al.(2020)Hu, Gauthier, Qian, Wilcox, and
  Levy}]{hu-etal-2020-systematic}
Jennifer Hu, Jon Gauthier, Peng Qian, Ethan Wilcox, and Roger Levy. 2020.
\newblock \href {https://doi.org/10.18653/v1/2020.acl-main.158} {A systematic
  assessment of syntactic generalization in neural language models}.
\newblock In \emph{Proceedings of the 58th Annual Meeting of the Association
  for Computational Linguistics}, pages 1725--1744, Online. Association for
  Computational Linguistics.

\bibitem[{Huang et~al.(2009)Huang, Li, and Li}]{2009chinesesyntax}
Cheng-Teh~James Huang, Yen-hui~Audrey Li, and Yafei Li. 2009.
\newblock \emph{The syntax of Chinese}, volume~10.
\newblock Cambridge University Press Cambridge.

\bibitem[{Huang et~al.(2018)Huang, Li, and Simpson}]{2018handbook}
C.T.~James Huang, Y.H.~Audrey Li, and Andrew Simpson. 2018.
\newblock \emph{The handbook of Chinese linguistics}.
\newblock John Wiley \& Sons.

\bibitem[{Kalouli et~al.(2023)Kalouli, Hu, Webb, Moss, and
  de~Paiva}]{kalouli2023}
Aikaterini-Lida Kalouli, Hai Hu, Alexander~F. Webb, Lawrence~S. Moss, and
  Valeria de~Paiva. 2023.
\newblock \href {https://doi.org/10.1162/coli_a_00465} {{Curing the SICK and
  Other NLI Maladies}}.
\newblock \emph{Computational Linguistics}, 49(1):199--243.

\bibitem[{Kaplan et~al.(2020)Kaplan, McCandlish, Henighan, Brown, Chess, Child,
  Gray, Radford, Wu, and Amodei}]{2020scaling}
Jared Kaplan, Sam McCandlish, Tom Henighan, Tom~B. Brown, Benjamin Chess, Rewon
  Child, Scott Gray, Alec Radford, Jeffrey Wu, and Dario Amodei. 2020.
\newblock \href {http://arxiv.org/abs/2001.08361} {Scaling laws for neural
  language models}.
\newblock \emph{CoRR}, abs/2001.08361.

\bibitem[{Liu(2013)}]{2013redred}
Chen-Sheng~Luther Liu. 2013.
\newblock \href {http://www.jstor.org/stable/42635248} {Reduplication of
  adjectives in chinese: a default state}.
\newblock \emph{Journal of East Asian Linguistics}, 22(2):101--132.

\bibitem[{Liu et~al.(2019)Liu, Ott, Goyal, Du, Joshi, Chen, Levy, Lewis,
  Zettlemoyer, and Stoyanov}]{2019RoBERTa}
Yinhan Liu, Myle Ott, Naman Goyal, Jingfei Du, Mandar Joshi, Danqi Chen, Omer
  Levy, Mike Lewis, Luke Zettlemoyer, and Veselin Stoyanov. 2019.
\newblock \href {http://arxiv.org/abs/1907.11692} {Roberta: {A} robustly
  optimized {BERT} pretraining approach}.
\newblock \emph{CoRR}, abs/1907.11692.

\bibitem[{Mikhailov et~al.(2022)Mikhailov, Shamardina, Ryabinin, Pestova,
  Smurov, and Artemova}]{rucola}
Vladislav Mikhailov, Tatiana Shamardina, Max Ryabinin, Alena Pestova, Ivan
  Smurov, and Ekaterina Artemova. 2022.
\newblock \href {https://doi.org/10.18653/v1/2022.emnlp-main.348}
  {{R}u{C}o{LA}: {R}ussian corpus of linguistic acceptability}.
\newblock In \emph{Proceedings of the 2022 Conference on Empirical Methods in
  Natural Language Processing}, pages 5207--5227, Abu Dhabi, United Arab
  Emirates. Association for Computational Linguistics.

\bibitem[{OpenAI(2023)}]{2023GPT4}
OpenAI. 2023.
\newblock \href {https://doi.org/10.48550/arXiv.2303.08774} {{GPT-4} technical
  report}.
\newblock \emph{CoRR}, abs/2303.08774.

\bibitem[{Ouyang et~al.(2022)Ouyang, Wu, Jiang, Almeida, Wainwright, Mishkin,
  Zhang, Agarwal, Slama, Ray, Schulman, Hilton, Kelton, Miller, Simens, Askell,
  Welinder, Christiano, Leike, and Lowe}]{2022InstructGPT}
Long Ouyang, Jeffrey Wu, Xu~Jiang, Diogo Almeida, Carroll~L. Wainwright, Pamela
  Mishkin, Chong Zhang, Sandhini Agarwal, Katarina Slama, Alex Ray, John
  Schulman, Jacob Hilton, Fraser Kelton, Luke Miller, Maddie Simens, Amanda
  Askell, Peter Welinder, Paul~F. Christiano, Jan Leike, and Ryan Lowe. 2022.
\newblock \href
  {http://papers.nips.cc/paper\_files/paper/2022/hash/b1efde53be364a73914f58805a001731-Abstract-Conference.html}
  {Training language models to follow instructions with human feedback}.
\newblock In \emph{NeurIPS}.

\bibitem[{Radford et~al.(2018)Radford, Narasimhan, Salimans, and
  Sutskever}]{2018GPT}
Alec Radford, Karthik Narasimhan, Tim Salimans, and Ilya Sutskever. 2018.
\newblock Improving language understanding by generative pre-training.

\bibitem[{Radford et~al.(2019)Radford, Wu, Child, Luan, Amodei, Sutskever
  et~al.}]{2019GPT2}
Alec Radford, Jeffrey Wu, Rewon Child, David Luan, Dario Amodei, Ilya
  Sutskever, et~al. 2019.
\newblock Language models are unsupervised multitask learners.
\newblock \emph{OpenAI blog}, 1(8):9.

\bibitem[{Rae et~al.(2021)Rae, Borgeaud, Cai, Millican, Hoffmann, Song,
  Aslanides, Henderson, Ring, Young, Rutherford, Hennigan, Menick, Cassirer,
  Powell, van~den Driessche, Hendricks, Rauh, Huang, Glaese, Welbl, Dathathri,
  Huang, Uesato, Mellor, Higgins, Creswell, McAleese, Wu, Elsen, Jayakumar,
  Buchatskaya, Budden, Sutherland, Simonyan, Paganini, Sifre, Martens, Li,
  Kuncoro, Nematzadeh, Gribovskaya, Donato, Lazaridou, Mensch, Lespiau,
  Tsimpoukelli, Grigorev, Fritz, Sottiaux, Pajarskas, Pohlen, Gong, Toyama,
  de~Masson~d'Autume, Li, Terzi, Mikulik, Babuschkin, Clark, de~Las~Casas, Guy,
  Jones, Bradbury, Johnson, Hechtman, Weidinger, Gabriel, Isaac, Lockhart,
  Osindero, Rimell, Dyer, Vinyals, Ayoub, Stanway, Bennett, Hassabis,
  Kavukcuoglu, and Irving}]{2021Gopher}
Jack~W. Rae, Sebastian Borgeaud, Trevor Cai, Katie Millican, Jordan Hoffmann,
  H.~Francis Song, John Aslanides, Sarah Henderson, Roman Ring, Susannah Young,
  Eliza Rutherford, Tom Hennigan, Jacob Menick, Albin Cassirer, Richard Powell,
  George van~den Driessche, Lisa~Anne Hendricks, Maribeth Rauh, Po{-}Sen Huang,
  Amelia Glaese, Johannes Welbl, Sumanth Dathathri, Saffron Huang, Jonathan
  Uesato, John Mellor, Irina Higgins, Antonia Creswell, Nat McAleese, Amy Wu,
  Erich Elsen, Siddhant~M. Jayakumar, Elena Buchatskaya, David Budden, Esme
  Sutherland, Karen Simonyan, Michela Paganini, Laurent Sifre, Lena Martens,
  Xiang~Lorraine Li, Adhiguna Kuncoro, Aida Nematzadeh, Elena Gribovskaya,
  Domenic Donato, Angeliki Lazaridou, Arthur Mensch, Jean{-}Baptiste Lespiau,
  Maria Tsimpoukelli, Nikolai Grigorev, Doug Fritz, Thibault Sottiaux, Mantas
  Pajarskas, Toby Pohlen, Zhitao Gong, Daniel Toyama, Cyprien
  de~Masson~d'Autume, Yujia Li, Tayfun Terzi, Vladimir Mikulik, Igor
  Babuschkin, Aidan Clark, Diego de~Las~Casas, Aurelia Guy, Chris Jones, James
  Bradbury, Matthew~J. Johnson, Blake~A. Hechtman, Laura Weidinger, Iason
  Gabriel, William~S. Isaac, Edward Lockhart, Simon Osindero, Laura Rimell,
  Chris Dyer, Oriol Vinyals, Kareem Ayoub, Jeff Stanway, Lorrayne Bennett,
  Demis Hassabis, Koray Kavukcuoglu, and Geoffrey Irving. 2021.
\newblock \href {http://arxiv.org/abs/2112.11446} {Scaling language models:
  Methods, analysis {\&} insights from training gopher}.
\newblock \emph{CoRR}, abs/2112.11446.

\bibitem[{Raffel et~al.(2020)Raffel, Shazeer, Roberts, Lee, Narang, Matena,
  Zhou, Li, and Liu}]{2019T5}
Colin Raffel, Noam Shazeer, Adam Roberts, Katherine Lee, Sharan Narang, Michael
  Matena, Yanqi Zhou, Wei Li, and Peter~J. Liu. 2020.
\newblock \href {http://jmlr.org/papers/v21/20-074.html} {Exploring the limits
  of transfer learning with a unified text-to-text transformer}.
\newblock \emph{J. Mach. Learn. Res.}, 21:140:1--140:67.

\bibitem[{Sanh et~al.(2022)Sanh, Webson, Raffel, Bach, Sutawika, Alyafeai,
  Chaffin, Stiegler, Raja, Dey, Bari, Xu, Thakker, Sharma, Szczechla, Kim,
  Chhablani, Nayak, Datta, Chang, Jiang, Wang, Manica, Shen, Yong, Pandey,
  Bawden, Wang, Neeraj, Rozen, Sharma, Santilli, F{\'{e}}vry, Fries, Teehan,
  Scao, Biderman, Gao, Wolf, and Rush}]{2021T0}
Victor Sanh, Albert Webson, Colin Raffel, Stephen~H. Bach, Lintang Sutawika,
  Zaid Alyafeai, Antoine Chaffin, Arnaud Stiegler, Arun Raja, Manan Dey,
  M~Saiful Bari, Canwen Xu, Urmish Thakker, Shanya~Sharma Sharma, Eliza
  Szczechla, Taewoon Kim, Gunjan Chhablani, Nihal~V. Nayak, Debajyoti Datta,
  Jonathan Chang, Mike~Tian{-}Jian Jiang, Han Wang, Matteo Manica, Sheng Shen,
  Zheng~Xin Yong, Harshit Pandey, Rachel Bawden, Thomas Wang, Trishala Neeraj,
  Jos Rozen, Abheesht Sharma, Andrea Santilli, Thibault F{\'{e}}vry, Jason~Alan
  Fries, Ryan Teehan, Teven~Le Scao, Stella Biderman, Leo Gao, Thomas Wolf, and
  Alexander~M. Rush. 2022.
\newblock \href {https://openreview.net/forum?id=9Vrb9D0WI4} {Multitask
  prompted training enables zero-shot task generalization}.
\newblock In \emph{The Tenth International Conference on Learning
  Representations, {ICLR} 2022, Virtual Event, April 25-29, 2022}.
  OpenReview.net.

\bibitem[{Scao et~al.(2022)Scao, Fan, Akiki, Pavlick, Ilic, Hesslow,
  Castagn{\'{e}}, Luccioni, Yvon, Gall{\'{e}}, Tow, Rush, Biderman, Webson,
  Ammanamanchi, Wang, Sagot, Muennighoff, del Moral, Ruwase, Bawden, Bekman,
  McMillan{-}Major, Beltagy, Nguyen, Saulnier, Tan, Suarez, Sanh,
  Lauren{\c{c}}on, Jernite, Launay, Mitchell, Raffel, Gokaslan, Simhi, Soroa,
  Aji, Alfassy, Rogers, Nitzav, Xu, Mou, Emezue, Klamm, Leong, van Strien,
  Adelani, and et~al.}]{2022BLOOM}
Teven~Le Scao, Angela Fan, Christopher Akiki, Ellie Pavlick, Suzana Ilic,
  Daniel Hesslow, Roman Castagn{\'{e}}, Alexandra~Sasha Luccioni,
  Fran{\c{c}}ois Yvon, Matthias Gall{\'{e}}, Jonathan Tow, Alexander~M. Rush,
  Stella Biderman, Albert Webson, Pawan~Sasanka Ammanamanchi, Thomas Wang,
  Beno{\^{\i}}t Sagot, Niklas Muennighoff, Albert~Villanova del Moral, Olatunji
  Ruwase, Rachel Bawden, Stas Bekman, Angelina McMillan{-}Major, Iz~Beltagy,
  Huu Nguyen, Lucile Saulnier, Samson Tan, Pedro~Ortiz Suarez, Victor Sanh,
  Hugo Lauren{\c{c}}on, Yacine Jernite, Julien Launay, Margaret Mitchell, Colin
  Raffel, Aaron Gokaslan, Adi Simhi, Aitor Soroa, Alham~Fikri Aji, Amit
  Alfassy, Anna Rogers, Ariel~Kreisberg Nitzav, Canwen Xu, Chenghao Mou, Chris
  Emezue, Christopher Klamm, Colin Leong, Daniel van Strien, David~Ifeoluwa
  Adelani, and et~al. 2022.
\newblock \href {https://doi.org/10.48550/arXiv.2211.05100} {{BLOOM:} {A}
  176b-parameter open-access multilingual language model}.
\newblock \emph{CoRR}, abs/2211.05100.

\bibitem[{Schütze(2016)}]{2016Schutze}
Carson~T. Schütze. 2016.
\newblock \href {https://doi.org/10.17169/langsci.b89.100} {\emph{The empirical
  base of linguistics}}.
\newblock Number~2 in Classics in Linguistics. Language Science Press, Berlin.

\bibitem[{Sinha et~al.(2023)Sinha, Gauthier, Mueller, Misra, Fuentes, Levy, and
  Williams}]{sinha-etal-2023-language}
Koustuv Sinha, Jon Gauthier, Aaron Mueller, Kanishka Misra, Keren Fuentes,
  Roger Levy, and Adina Williams. 2023.
\newblock \href {https://doi.org/10.18653/v1/2023.acl-long.333} {Language model
  acceptability judgements are not always robust to context}.
\newblock In \emph{Proceedings of the 61st Annual Meeting of the Association
  for Computational Linguistics (Volume 1: Long Papers)}, pages 6043--6063,
  Toronto, Canada. Association for Computational Linguistics.

\bibitem[{Smith et~al.(2022)Smith, Patwary, Norick, LeGresley, Rajbhandari,
  Casper, Liu, Prabhumoye, Zerveas, Korthikanti, Zheng, Child, Aminabadi,
  Bernauer, Song, Shoeybi, He, Houston, Tiwary, and Catanzaro}]{2022MT-NLG}
Shaden Smith, Mostofa Patwary, Brandon Norick, Patrick LeGresley, Samyam
  Rajbhandari, Jared Casper, Zhun Liu, Shrimai Prabhumoye, George Zerveas,
  Vijay Korthikanti, Elton Zheng, Rewon Child, Reza~Yazdani Aminabadi, Julie
  Bernauer, Xia Song, Mohammad Shoeybi, Yuxiong He, Michael Houston, Saurabh
  Tiwary, and Bryan Catanzaro. 2022.
\newblock \href {http://arxiv.org/abs/2201.11990} {Using deepspeed and megatron
  to train megatron-turing {NLG} 530b, {A} large-scale generative language
  model}.
\newblock \emph{CoRR}, abs/2201.11990.

\bibitem[{Swayamdipta et~al.(2020)Swayamdipta, Schwartz, Lourie, Wang,
  Hajishirzi, Smith, and Choi}]{swayamdipta-etal-2020-dataset-cartography}
Swabha Swayamdipta, Roy Schwartz, Nicholas Lourie, Yizhong Wang, Hannaneh
  Hajishirzi, Noah~A. Smith, and Yejin Choi. 2020.
\newblock \href {https://doi.org/10.18653/v1/2020.emnlp-main.746} {Dataset
  cartography: Mapping and diagnosing datasets with training dynamics}.
\newblock In \emph{Proceedings of the 2020 Conference on Empirical Methods in
  Natural Language Processing (EMNLP)}, pages 9275--9293, Online. Association
  for Computational Linguistics.

\bibitem[{Tenney et~al.(2019)Tenney, Das, and Pavlick}]{2019pipeline}
Ian Tenney, Dipanjan Das, and Ellie Pavlick. 2019.
\newblock \href {https://doi.org/10.18653/v1/p19-1452} {{BERT} rediscovers the
  classical {NLP} pipeline}.
\newblock In \emph{Proceedings of the 57th Conference of the Association for
  Computational Linguistics, {ACL} 2019, Florence, Italy, July 28- August 2,
  2019, Volume 1: Long Papers}, pages 4593--4601. Association for Computational
  Linguistics.

\bibitem[{Thoppilan et~al.(2022)Thoppilan, Freitas, Hall, Shazeer,
  Kulshreshtha, Cheng, Jin, Bos, Baker, Du, Li, Lee, Zheng, Ghafouri, Menegali,
  Huang, Krikun, Lepikhin, Qin, Chen, Xu, Chen, Roberts, Bosma, Zhou, Chang,
  Krivokon, Rusch, Pickett, Meier{-}Hellstern, Morris, Doshi, Santos, Duke,
  Soraker, Zevenbergen, Prabhakaran, Diaz, Hutchinson, Olson, Molina,
  Hoffman{-}John, Lee, Aroyo, Rajakumar, Butryna, Lamm, Kuzmina, Fenton, Cohen,
  Bernstein, Kurzweil, Aguera{-}Arcas, Cui, Croak, Chi, and Le}]{2022LaMDA}
Romal Thoppilan, Daniel~De Freitas, Jamie Hall, Noam Shazeer, Apoorv
  Kulshreshtha, Heng{-}Tze Cheng, Alicia Jin, Taylor Bos, Leslie Baker, Yu~Du,
  YaGuang Li, Hongrae Lee, Huaixiu~Steven Zheng, Amin Ghafouri, Marcelo
  Menegali, Yanping Huang, Maxim Krikun, Dmitry Lepikhin, James Qin, Dehao
  Chen, Yuanzhong Xu, Zhifeng Chen, Adam Roberts, Maarten Bosma, Yanqi Zhou,
  Chung{-}Ching Chang, Igor Krivokon, Will Rusch, Marc Pickett, Kathleen~S.
  Meier{-}Hellstern, Meredith~Ringel Morris, Tulsee Doshi, Renelito~Delos
  Santos, Toju Duke, Johnny Soraker, Ben Zevenbergen, Vinodkumar Prabhakaran,
  Mark Diaz, Ben Hutchinson, Kristen Olson, Alejandra Molina, Erin
  Hoffman{-}John, Josh Lee, Lora Aroyo, Ravi Rajakumar, Alena Butryna, Matthew
  Lamm, Viktoriya Kuzmina, Joe Fenton, Aaron Cohen, Rachel Bernstein, Ray
  Kurzweil, Blaise Aguera{-}Arcas, Claire Cui, Marian Croak, Ed~H. Chi, and
  Quoc Le. 2022.
\newblock \href {http://arxiv.org/abs/2201.08239} {Lamda: Language models for
  dialog applications}.
\newblock \emph{CoRR}, abs/2201.08239.

\bibitem[{Trotta et~al.(2021)Trotta, Guarasci, Leonardelli, and
  Tonelli}]{itacola}
Daniela Trotta, Raffaele Guarasci, Elisa Leonardelli, and Sara Tonelli. 2021.
\newblock \href {https://doi.org/10.18653/v1/2021.findings-emnlp.250}
  {Monolingual and cross-lingual acceptability judgments with the {I}talian
  {C}o{LA} corpus}.
\newblock In \emph{Findings of the Association for Computational Linguistics:
  EMNLP 2021}, pages 2929--2940, Punta Cana, Dominican Republic. Association
  for Computational Linguistics.

\bibitem[{Vaswani et~al.(2017)Vaswani, Shazeer, Parmar, Uszkoreit, Jones,
  Gomez, Kaiser, and Polosukhin}]{2017Transformer}
Ashish Vaswani, Noam Shazeer, Niki Parmar, Jakob Uszkoreit, Llion Jones,
  Aidan~N. Gomez, Lukasz Kaiser, and Illia Polosukhin. 2017.
\newblock \href
  {https://proceedings.neurips.cc/paper/2017/hash/3f5ee243547dee91fbd053c1c4a845aa-Abstract.html}
  {Attention is all you need}.
\newblock In \emph{Advances in Neural Information Processing Systems 30: Annual
  Conference on Neural Information Processing Systems 2017, December 4-9, 2017,
  Long Beach, CA, {USA}}, pages 5998--6008.

\bibitem[{Wang et~al.(2019)Wang, Pruksachatkun, Nangia, Singh, Michael, Hill,
  Levy, and Bowman}]{2019SGLUE}
Alex Wang, Yada Pruksachatkun, Nikita Nangia, Amanpreet Singh, Julian Michael,
  Felix Hill, Omer Levy, and Samuel~R. Bowman. 2019.
\newblock \href
  {https://proceedings.neurips.cc/paper/2019/hash/4496bf24afe7fab6f046bf4923da8de6-Abstract.html}
  {Superglue: {A} stickier benchmark for general-purpose language understanding
  systems}.
\newblock In \emph{Advances in Neural Information Processing Systems 32: Annual
  Conference on Neural Information Processing Systems 2019, NeurIPS 2019,
  December 8-14, 2019, Vancouver, BC, Canada}, pages 3261--3275.

\bibitem[{Wang et~al.(2018)Wang, Singh, Michael, Hill, Levy, and
  Bowman}]{2018GLUE}
Alex Wang, Amanpreet Singh, Julian Michael, Felix Hill, Omer Levy, and
  Samuel~R. Bowman. 2018.
\newblock \href {https://doi.org/10.18653/v1/w18-5446} {{GLUE:} {A} multi-task
  benchmark and analysis platform for natural language understanding}.
\newblock In \emph{Proceedings of the Workshop: Analyzing and Interpreting
  Neural Networks for NLP, BlackboxNLP@EMNLP 2018, Brussels, Belgium, November
  1, 2018}, pages 353--355. Association for Computational Linguistics.

\bibitem[{Wang et~al.(2022)Wang, Mishra, Alipoormolabashi, Kordi, Mirzaei,
  Naik, Ashok, Dhanasekaran, Arunkumar, Stap, Pathak, Karamanolakis, Lai,
  Purohit, Mondal, Anderson, Kuznia, Doshi, Pal, Patel, Moradshahi, Parmar,
  Purohit, Varshney, Kaza, Verma, Puri, Karia, Doshi, Sampat, Mishra, A, Patro,
  Dixit, and Shen}]{2022SuperNatural}
Yizhong Wang, Swaroop Mishra, Pegah Alipoormolabashi, Yeganeh Kordi, Amirreza
  Mirzaei, Atharva Naik, Arjun Ashok, Arut~Selvan Dhanasekaran, Anjana
  Arunkumar, David Stap, Eshaan Pathak, Giannis Karamanolakis, Haizhi~Gary Lai,
  Ishan Purohit, Ishani Mondal, Jacob Anderson, Kirby Kuznia, Krima Doshi,
  Kuntal~Kumar Pal, Maitreya Patel, Mehrad Moradshahi, Mihir Parmar, Mirali
  Purohit, Neeraj Varshney, Phani~Rohitha Kaza, Pulkit Verma, Ravsehaj~Singh
  Puri, Rushang Karia, Savan Doshi, Shailaja~Keyur Sampat, Siddhartha Mishra,
  Sujan~Reddy A, Sumanta Patro, Tanay Dixit, and Xudong Shen. 2022.
\newblock \href {https://aclanthology.org/2022.emnlp-main.340}
  {Super-naturalinstructions: Generalization via declarative instructions on
  1600+ {NLP} tasks}.
\newblock In \emph{Proceedings of the 2022 Conference on Empirical Methods in
  Natural Language Processing, {EMNLP} 2022, Abu Dhabi, United Arab Emirates,
  December 7-11, 2022}, pages 5085--5109. Association for Computational
  Linguistics.

\bibitem[{Warstadt et~al.(2020)Warstadt, Parrish, Liu, Mohananey, Peng, Wang,
  and Bowman}]{2019BLiMP}
Alex Warstadt, Alicia Parrish, Haokun Liu, Anhad Mohananey, Wei Peng,
  Sheng{-}Fu Wang, and Samuel~R. Bowman. 2020.
\newblock \href {https://doi.org/10.1162/tacl\_a\_00321} {Blimp: The benchmark
  of linguistic minimal pairs for english}.
\newblock \emph{Trans. Assoc. Comput. Linguistics}, 8:377--392.

\bibitem[{Warstadt et~al.(2019)Warstadt, Singh, and Bowman}]{2018CoLA}
Alex Warstadt, Amanpreet Singh, and Samuel~R. Bowman. 2019.
\newblock \href {https://doi.org/10.1162/tacl\_a\_00290} {Neural network
  acceptability judgments}.
\newblock \emph{Trans. Assoc. Comput. Linguistics}, 7:625--641.

\bibitem[{Wei et~al.(2022{\natexlab{a}})Wei, Bosma, Zhao, Guu, Yu, Lester, Du,
  Dai, and Le}]{2021FLAN}
Jason Wei, Maarten Bosma, Vincent~Y. Zhao, Kelvin Guu, Adams~Wei Yu, Brian
  Lester, Nan Du, Andrew~M. Dai, and Quoc~V. Le. 2022{\natexlab{a}}.
\newblock \href {https://openreview.net/forum?id=gEZrGCozdqR} {Finetuned
  language models are zero-shot learners}.
\newblock In \emph{The Tenth International Conference on Learning
  Representations, {ICLR} 2022, Virtual Event, April 25-29, 2022}.
  OpenReview.net.

\bibitem[{Wei et~al.(2022{\natexlab{b}})Wei, Tay, Bommasani, Raffel, Zoph,
  Borgeaud, Yogatama, Bosma, Zhou, Metzler, Chi, Hashimoto, Vinyals, Liang,
  Dean, and Fedus}]{2022emergent}
Jason Wei, Yi~Tay, Rishi Bommasani, Colin Raffel, Barret Zoph, Sebastian
  Borgeaud, Dani Yogatama, Maarten Bosma, Denny Zhou, Donald Metzler, Ed~H.
  Chi, Tatsunori Hashimoto, Oriol Vinyals, Percy Liang, Jeff Dean, and William
  Fedus. 2022{\natexlab{b}}.
\newblock \href {https://openreview.net/forum?id=yzkSU5zdwD} {Emergent
  abilities of large language models}.
\newblock \emph{Trans. Mach. Learn. Res.}, 2022.

\bibitem[{Wolf et~al.(2020)Wolf, Debut, Sanh, Chaumond, Delangue, Moi, Cistac,
  Rault, Louf, Funtowicz, Davison, Shleifer, von Platen, Ma, Jernite, Plu, Xu,
  Scao, Gugger, Drame, Lhoest, and Rush}]{2020transformers}
Thomas Wolf, Lysandre Debut, Victor Sanh, Julien Chaumond, Clement Delangue,
  Anthony Moi, Pierric Cistac, Tim Rault, Rémi Louf, Morgan Funtowicz, Joe
  Davison, Sam Shleifer, Patrick von Platen, Clara Ma, Yacine Jernite, Julien
  Plu, Canwen Xu, Teven~Le Scao, Sylvain Gugger, Mariama Drame, Quentin Lhoest,
  and Alexander~M. Rush. 2020.
\newblock \href {https://www.aclweb.org/anthology/2020.emnlp-demos.6}
  {Transformers: State-of-the-art natural language processing}.
\newblock In \emph{Proceedings of the 2020 Conference on Empirical Methods in
  Natural Language Processing: System Demonstrations}, pages 38--45, Online.
  Association for Computational Linguistics.

\bibitem[{Xiang et~al.(2021)Xiang, Yang, Li, Warstadt, and Kann}]{2021CLiMP}
Beilei Xiang, Changbing Yang, Yu~Li, Alex Warstadt, and Katharina Kann. 2021.
\newblock \href {https://doi.org/10.18653/v1/2021.eacl-main.242} {Climp: {A}
  benchmark for chinese language model evaluation}.
\newblock In \emph{Proceedings of the 16th Conference of the European Chapter
  of the Association for Computational Linguistics: Main Volume, {EACL} 2021,
  Online, April 19 - 23, 2021}, pages 2784--2790. Association for Computational
  Linguistics.

\bibitem[{Xue et~al.(2021)Xue, Constant, Roberts, Kale, Al{-}Rfou, Siddhant,
  Barua, and Raffel}]{2020mT5}
Linting Xue, Noah Constant, Adam Roberts, Mihir Kale, Rami Al{-}Rfou, Aditya
  Siddhant, Aditya Barua, and Colin Raffel. 2021.
\newblock \href {https://doi.org/10.18653/v1/2021.naacl-main.41} {mt5: {A}
  massively multilingual pre-trained text-to-text transformer}.
\newblock In \emph{Proceedings of the 2021 Conference of the North American
  Chapter of the Association for Computational Linguistics: Human Language
  Technologies, {NAACL-HLT} 2021, Online, June 6-11, 2021}, pages 483--498.
  Association for Computational Linguistics.

\bibitem[{Yang et~al.(2019)Yang, Dai, Yang, Carbonell, Salakhutdinov, and
  Le}]{2019XLNet}
Zhilin Yang, Zihang Dai, Yiming Yang, Jaime~G. Carbonell, Ruslan Salakhutdinov,
  and Quoc~V. Le. 2019.
\newblock \href
  {https://proceedings.neurips.cc/paper/2019/hash/dc6a7e655d7e5840e66733e9ee67cc69-Abstract.html}
  {Xlnet: Generalized autoregressive pretraining for language understanding}.
\newblock In \emph{Advances in Neural Information Processing Systems 32: Annual
  Conference on Neural Information Processing Systems 2019, NeurIPS 2019,
  December 8-14, 2019, Vancouver, BC, Canada}, pages 5754--5764.

\bibitem[{Yao et~al.(2018)Yao, Xie, Lin, and Huang}]{yao2018}
Yao Yao, Zhiguo Xie, Chien-Jer~Charles Lin, and Chu-Ren Huang. 2018.
\newblock Acceptability or grammaticality: Judging chinese sentences for
  language studies.
\newblock In Chu-Ren Huang, Yen-Hwei Lin, and I-Hsuan Chen, editors,
  \emph{Cambridge Handbook of Chinese Linguistics}.

\bibitem[{Zhang et~al.(2022)Zhang, Roller, Goyal, Artetxe, Chen, Chen, Dewan,
  Diab, Li, Lin, Mihaylov, Ott, Shleifer, Shuster, Simig, Koura, Sridhar, Wang,
  and Zettlemoyer}]{2022OPT}
Susan Zhang, Stephen Roller, Naman Goyal, Mikel Artetxe, Moya Chen, Shuohui
  Chen, Christopher Dewan, Mona~T. Diab, Xian Li, Xi~Victoria Lin, Todor
  Mihaylov, Myle Ott, Sam Shleifer, Kurt Shuster, Daniel Simig, Punit~Singh
  Koura, Anjali Sridhar, Tianlu Wang, and Luke Zettlemoyer. 2022.
\newblock \href {https://doi.org/10.48550/arXiv.2205.01068} {{OPT:} open
  pre-trained transformer language models}.
\newblock \emph{CoRR}, abs/2205.01068.

\end{thebibliography}
\bibliographystyle{acl_natbib}

\newpage
\appendix
\section{Training Details}\label{sec:training-detail}
The training hyper-parameters are given in Table~\ref{tab:hyperparam}. Our code is implemented with Transformers~\citep{2020transformers} library version 4.25.1, and the training is conducted on an RTX 3090 with 24GB RAM. The search space is $\{$5e-6, 7.5e-6, 1e-5$\}$ for learning rate, and $\{$0.01, 0.05, 0.1$\}$ for weight decay.

When training on down-sampled data (Figure~\ref{fig:transfer}), all models are trained for one thousand steps with other hyper-parameters unchanged.
\begin{table}[h]
    \centering
    \begin{tabular}{m{3cm}cc}
    \toprule
       \textbf{Hyperparam}  & \textbf{fine-tune} & \textbf{prompt-tune} \\\midrule
       Learning Rate & 7.5e-6 & 1e-5 \\
       Batch Size & 32 & 32 \\
       Weight Decay & 0.05 & 0.01 \\
       Epochs & 15 & 10\\
       LR Decay & linear & linear \\
       Warmup Ratio & 0.06 & 0.06 \\
    \bottomrule
    \end{tabular}
    \caption{Hyper-parameters for training XLM-RoBERTa.}
    \label{tab:hyperparam}
\end{table}

\section{Prompts}\label{sec:prompt}
For InstructGPT, we use OpenAI's text-ada-001, text-babbage-001, text-curie-001, and text-davinci-001 APIs, which presumably correspond to the 350M, 1.3B, 6.7B, and 175B versions of InstructGPT. For ChatGPT, we use OpenAI's gpt-3.5-turbo API, the size of which remains unknown to the public. All models use greedy decoding to generate the answer. These models are accessed on May 21, 2023.

For mTk-Instruct, we adopt the original instruction for CoLA in NIV2 task 616\footnote{\url{https://github.com/allenai/natural-instructions/blob/master/tasks/task616_cola_classification.json}}, but replace the in-context examples with the Chinese ones in Table~\ref{tab:prompts}.


\begin{table}[t]
    \centering
    \scalebox{0.9}{
    \begin{tabular}{|l|}
    \toprule
    Our friends won't buy this analysis, let alone the\\
    next one we propose.\\
    Question: acceptable or unacceptable? acceptable\\
    \\
    They drank the pub.\\
    Question: acceptable or unacceptable? unacceptable\\
    \\
    We yelled ourselves hoarse.\\
    Question: acceptable or unacceptable? acceptable\\
    \\
    Harry coughed himself.\\
    Question: acceptable or unacceptable? unacceptable\\
    \\
    Bill followed the road into the forest.\\
    Question: acceptable or unacceptable? acceptable\\
    \\
    \{sent\}\\
    Question: acceptable or unacceptable? \\
    \bottomrule
    \end{tabular}}
    \caption{Prompts for InstructGPT and ChatGPT on CoLA.}
    \label{tab:prompts-cola}
\end{table}
\begin{CJK*}{UTF8}{gkai}
\begin{table}[t]
    \centering
    \scalebox{0.9}{
    \begin{tabular}{|l|}
    \toprule
    玛丽的儿子考上了北大。\\
    Question: acceptable or unacceptable? acceptable\\
    \\
    张三被李四打了自己。\\
    Question: acceptable or unacceptable? unacceptable\\
    \\
    张三的房间还好。\\
    Question: acceptable or unacceptable? acceptable\\
    \\
    宿舍被他买花去了。\\
    Question: acceptable or unacceptable? unacceptable\\
    \\
    一个人很聪明。\\
    Question: acceptable or unacceptable? acceptable\\
    \\
    \{sent\}\\
    Question: acceptable or unacceptable? \\
    \bottomrule
    \end{tabular}}
    \caption{Prompts for InstructGPT and ChatGPT on CoLAC.}
    \label{tab:prompts}
\end{table}
\end{CJK*}

\section{Label Descriptions}\label{sec:label}
In Table~\ref{tab:labels}, we provide a short explanation and an illustrative pair of sentences for the 18 syntaxtic phenomena in CoLAC.

\begin{CJK*}{UTF8}{gkai}
\begin{table*}[t]
    \centering
    \scalebox{.95}{
    \begin{tabular}{cll}
    \toprule
    \multirow{2}{*}{Label Type}         &     \multirow{2}{*}{Definition}      & Acceptable Example/ \\
    &&Unacceptable Example  \\ \midrule
    \multirow{2}{*}{\textsc{Argument structure}}&\multirow{2}{6.5cm}{the argument-taking ability of a predicate} & 他走了。               \\&& 他走了我们。               \\ \cmidrule{2-3}
    \multirow{2}{*}{\textsc{Aspect}}            &\multirow{2}{6.5cm}{a category that expresses the temporal structure of an eventuality}& 警察踹门进屋搜寻。          \\&& 警察踹着门进屋搜寻。           \\ \cmidrule{2-3}
    \multirow{2}{*}{\textsc{BA-construction}}   &\multirow{2}{6.5cm}{a Chinese-specific construction involving the functional word BA} & 我会把他恨一辈子。          \\&& 我会把他怕一辈子。            \\ \cmidrule{2-3}
    \multirow{2}{*}{\textsc{Binding}}           &\multirow{2}{6.5cm}{the referential possibility of a noun phrase as constrained by structure} & 玛丽常常流眼泪。           \\&& 玛丽常常流自己的眼泪。          \\ \cmidrule{2-3}
    \multirow{3}{*}{\textsc{Case}}              &\multirow{3}{6.5cm}{the requirement that nominal arguments be licensed by certain lexical categories (which typically refer to Vs and Ps)} & \multirow{3}{4.5cm}{我对他的去世非常伤心。我非常伤心他的去世。}      \\            \\
    \\ \cmidrule{2-3}
    \multirow{2}{*}{\textsc{Compounding}}       &\multirow{2}{6.5cm}{a morphologically complex word built from more than one root} & 奶奶织大了毛衣。           \\&& 奶奶织太大了毛衣。            \\ \cmidrule{2-3}
    \multirow{2}{*}{\textsc{Conjunction}}       &\multirow{2}{6.5cm}{a category that links two or more syntactic units of various kinds} & 我去，或是你来。           \\&& 我去，你或是来。             \\ \cmidrule{2-3}
    \multirow{2}{*}{\textsc{Ellipsis}}          &\multirow{2}{6.5cm}{the phenomenon of meaning without form, e.g., sluicing and null arguments} & 他不喜欢张三，我也不喜欢。   \\&   & 他不喜欢张三，我也不是。         \\ \cmidrule{2-3}
    \multirow{2}{*}{\textsc{Modal}}             &\multirow{2}{6.5cm}{a category that expresses permission, possibility, obligation and so on} & 这首小曲儿可能会长一点。     \\&  & 这首小曲儿能会长一点。          \\ \cmidrule{2-3}
    \multirow{3}{*}{\textsc{Movement}}          &\multirow{3}{6.5cm}{the pronunciation of a phrase in a position different from where the phrase is interpreted} & \multirow{3}{4.5cm}{书，你认为他看完了吗？你书认为他看完了吗？}       \\               \\
    \\ \cmidrule{2-3}
    \multirow{2}{*}{\textsc{Noun phrase}}       &\multirow{2}{6.5cm}{data with this label bear on the internal structure of the noun phrase} & 我对他们三个特别好。        \\& & 我对三个他们特别好。           \\ \cmidrule{2-3}
    \multirow{3}{*}{\textsc{Passive}}           &\multirow{3}{6.5cm}{a construction where a less thematically prominent noun phrase serves as the surface subject } & \multirow{3}{4.7cm}{张三被李四在学校骗走了。张三被在学校骗走了。}      \\            \\
    \\ \cmidrule{2-3}
    \multirow{2}{*}{\textsc{Phrase structure}}  &\multirow{2}{6.5cm}{structural restrictions on the ordering of different sub-constituents in a sentence} & 他大声唱着歌。            \\&& 他唱着大声歌。              \\ \cmidrule{2-3}
    \multirow{2}{*}{\textsc{Question}}          &\multirow{2}{6.5cm}{a sentence type that typically seeks information from the hearer(s)} & 张三买书不买书？           \\&& 张三买书张三不买书？           \\ \cmidrule{2-3}
    \multirow{2}{*}{\textsc{Reduplication}}     &\multirow{2}{6.5cm}{a morphological process where a certain part of a word is repeated} & 他应该明白明白这个道理。     \\&  & 他应该明明白白这个道理。         \\ \cmidrule{2-3}
    \multirow{2}{*}{\textsc{Relative clause}}   &\multirow{2}{6.5cm}{a clause that modifies a noun phrase} & 李四看到张三的时候         \\& & 李四所看到张三的时候           \\ \cmidrule{2-3}
    \multirow{2}{*}{\textsc{Resumptive}}        &\multirow{2}{6.5cm}{the occurrence of a pronoun in a structural position which forbids movement} & 那个人，我无法跟他合作。    \\&   & 那个人，我无法跟合作。          \\ \cmidrule{2-3}
    \multirow{2}{*}{\textsc{Topic construction}}&\multirow{2}{6.5cm}{a construction where a topic is followed by a clause that serves as its comment} & 学生，我以为吃了蛋糕。      \\&  & 三个学生，我以为吃了蛋糕。        \\ \bottomrule
    \end{tabular}}
    \caption{Example pairs of the 18 syntactic phenomena.}
    \label{tab:labels}
\end{table*}
\end{CJK*}

\end{document}